\documentclass[conference]{IEEEtran}
\IEEEoverridecommandlockouts
\usepackage{cite}
\usepackage{amsmath,amssymb,amsfonts}
\usepackage{algorithmic}
\usepackage{graphicx}
\usepackage{textcomp}
\usepackage{graphicx}
\usepackage{amsmath}
\usepackage{amssymb}
\usepackage{booktabs}
\usepackage{multirow}
\usepackage{bbding}
\usepackage{hyperref}
\usepackage{tabularx}

\usepackage{bibspacing}
\usepackage{caption}
\usepackage{subcaption}
\usepackage{makecell}
\usepackage{xcolor}
\def\BibTeX{{\rm B\kern-.05em{\sc i\kern-.025em b}\kern-.08em
    T\kern-.1667em\lower.7ex\hbox{E}\kern-.125emX}}
\begin{document}

\title{Uncertainty-Guided Appearance-Motion Association Network for Out-of-Distribution Action Detection\\
\thanks{Corresponding author: Xiang Fang. This research is part of the programme DesCartes and is supported by the National Research Foundation, Prime Minister’s Office, Singapore under its Campus for Research Excellence and Technological Enterprise (CREATE) programme.}}

\author{\IEEEauthorblockN{Xiang Fang}
\IEEEauthorblockA{\textit{Interdisciplinary Graduate Programme ERI@N} \\
\textit{College of Computing and Data Science} \\
\textit{Nanyang Technological University, Singapore}\\
\textit{CNRS@CREATE} \\
Singapore \\
xiang003@e.ntu.edu.sg}
\and
\IEEEauthorblockN{Arvind Easwaran}
\IEEEauthorblockA{\textit{College of Computing and Data Science} \\
\textit{Nanyang Technological University, Singapore}\\
Singapore \\
arvinde@ntu.edu.sg}
\and
\IEEEauthorblockN{Blaise Genest}
\IEEEauthorblockA{\textit{CNRS@CREATE} \\
Singapore\\
blaise.genest@cnrsatcreate.sg}
}

\maketitle

\begin{abstract}
Out-of-distribution (OOD) detection targets to detect and reject test samples with semantic shifts, to prevent models trained on in-distribution (ID) dataset from producing unreliable predictions. Existing works only extract the appearance features on image datasets, and cannot handle dynamic multimedia scenarios with much motion information. Therefore, we target a more realistic and challenging OOD detection task:
OOD action detection (ODAD). Given an untrimmed video,  ODAD  first classifies the ID actions and recognizes the OOD actions, and then localizes ID and OOD actions.  To this end, in this paper, we propose a novel Uncertainty-Guided Appearance-Motion Association Network (UAAN), which explores both appearance features and motion contexts to 
reason spatial-temporal inter-object interaction for ODAD.
Firstly, we  design separate appearance and motion   branches to extract corresponding appearance-oriented and motion-aspect object representations. In each branch, we construct a spatial-temporal graph to reason appearance-guided and  motion-driven  inter-object interaction. 
Then, we design an appearance-motion attention module to fuse the appearance and motion features for final action detection. Experimental results on two challenging datasets show that  UAAN beats state-of-the-art methods by a significant margin, illustrating its effectiveness.
\end{abstract}
\begin{IEEEkeywords}
uncertainty-guided appearance-motion association, out-of-distribution action detection
\end{IEEEkeywords}

\begin{figure}[t!]
    \centering
    \includegraphics[width=0.48\textwidth]{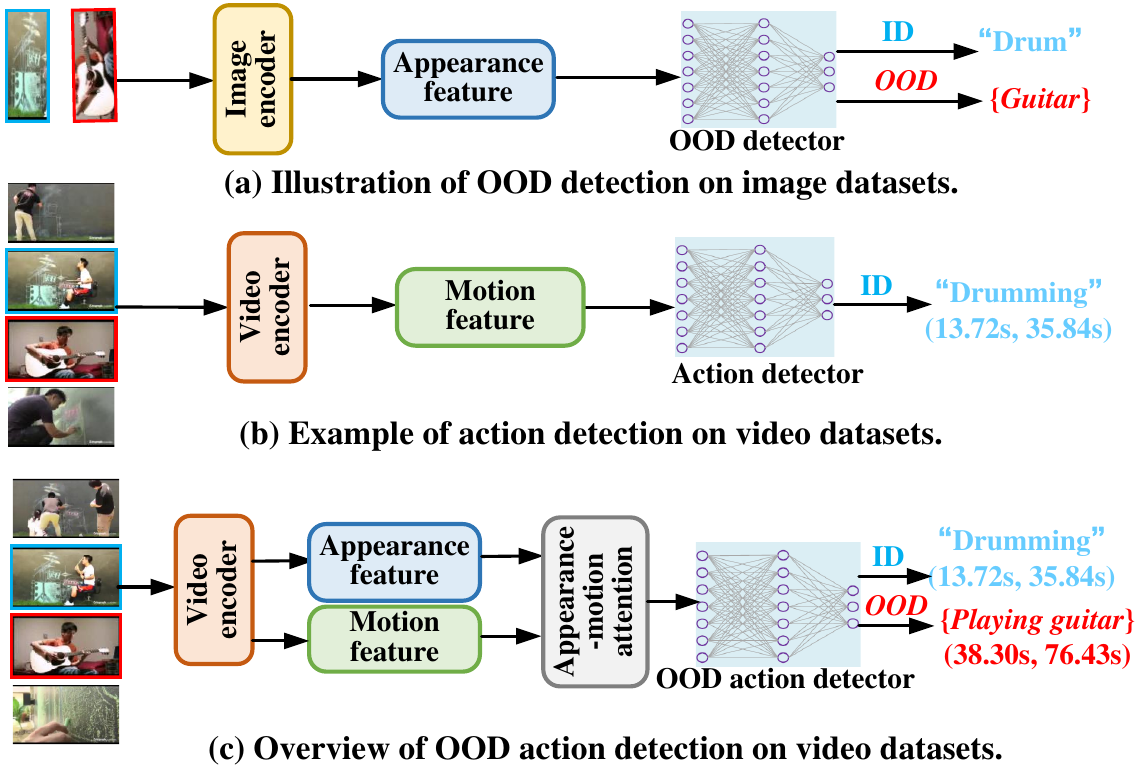}
    \caption{(a) Illustration of out-of-distribution (OOD)  detection that only detects  static images. (b) Previous temporal action detection models only classify/localize the ID actions and cannot detect OOD actions. (c) Our target task: OOD action detection that can not only classify and localize ID actions, but also detect and localize OOD actions. For the boxes and text, we color ID actions as {\color{blue}blue} and  OOD actions as {\color{red}red}, and background is not colored. The output labels on OOD actions are only for illustration, and they are not the actual output of the model. Best viewed in color.}
    \label{fig:intro}
\end{figure}

\section{Introduction}
\label{sec:intro}
Deep neural networks (DNNs) \cite{sun2023ai,tian2020deep,chen2021explainable,clement2020interactive,cheng2022attentive,roy2021image,li2023learning,robinson2023balancing}  have achieved impressive  success on the multimedia sample detection task under a closed-set assumption \cite{huo2022action,long2022retrieval,niu2009towards,baba2023impact,koneripalli2020rate,nakamura2018encryption,zhao2023open,yang2021background,hu2023learning}, where all the classes experienced during the test have been seen during training \cite{yanai2015food,tsukuda2023and,kastner2020estimating,queluz1999content,ruan2023video,xie2022mmeat,liu2023exploring,wang2025taylor,fang2026towardsicml,kuai2026dynamic,wang2025point,fang2025your,zhang2025monoattack,fang2023hierarchical,liu2024towards,yang2025eood,fang2022multi,fang2026cogniVerse,lei2025exploring,fang2023you,wang2025dypolyseg,fang2025hierarchical,yan2026fit,fang2025adaptive,wang2026topadapter,cai2025imperceptible,fang2026slap,wang2026reasoning,fang2026immuno,wang2026biologically,fang2026disentangling,wang2025reducing,fang2026advancing,fang2026unveiling,wang2026from,liu2023conditional,liu2026attacking,fang2026rethinking,wang2025seeing,fang2026towards,fang2025multi,fang2024fewer,liu2024pandora,fang2024multi,fang2025turing,fang2024not,liu2023hypotheses,fang2024rethinking,liu2024unsupervised,fang2023annotations,xiong2024rethinking,fang2021unbalanced,wang2025prototype,zhang2025manipulating,fang2026align,tang2024reparameterization,fang2025adaptivetai,tang2025simplification,fang2021animc,cai2026towards,fang2020v}. 
However, vanilla DNN-based methods compulsorily classify each sample into some of the known classes \cite{huang2023tracking,atrey2010multimodal,aygun2004modeling,wang2021feature,luo2019taking,wu2013using,si2019exploration,ouyang2019attentiondrop,xu2023mmlock}. It will lead to irrecoverable losses when classifying some outliers into wrong classes in some high-risk scenarios, such as autonomous driving \cite{zendel2022unifying,li2023domain,kurihata2005rainy,wang2016pathmon,patil2018geosclean,li2017compactness,ju2022fusing}. 
Thus, the out-of-distribution (OOD) detection task \cite{olber2023detection,hsu2020generalized,zisselman2020deep,wang2022vim} is proposed to accurately detect these  outliers from OOD classes and correctly classify  the samples from in-distribution (ID) classes during testing. 
Although most OOD detection methods \cite{liang2018enhancing,liu2020energy,sun2021react,lee2018simple,mohseni2020self,ming2023exploit} have achieved remarkable performance, they only focus on static images shown in Fig. \ref{fig:intro}, which limits their multimedia applications, where multimedia data is dynamically varying and not static. For example, in the  autonomous driving scenario, we have to respond when an unknown object appears or moves on the road.
These autonomous driving videos are dynamic, while previous OOD detection works focus on image datasets under the static assumption.
It is unrealistic to first divide each video into multiple images and then detect unknown objects/actions \cite{garg2013variational,iqbal2008compressed,yan2018participation,zhang2018systematic,tonomura1994structured}. This dynamic OOD detection problem based on videos is largely unexplored in the studies.

To handle dynamic OOD detection, we consider a challenging task: out-of-distribution action detection (ODAD)  shown in  Fig. \ref{fig:intro}. Given an untrimmed video, ODAD aims to classify ID actions, recognize OOD actions, and localize both ID and OOD actions. Although many action detection methods have been proposed to classify ID actions in an untrimmed video, few methods can correctly recognize OOD actions. An intuitive idea is to \textit{combine the OOD detection task and the action detection task to solve the more challenging ODAD task}. We observe that there are some gaps between the two topics: 1) data  gap, the OOD detection task focuses on images, while the action detection task targets videos; 2) feature  gap, the OOD detection task extracts appearance features, while the action detection mainly extracts motion features; 3) output gap, the OOD detection task does not conduct localization, while the action detection task needs to localize the target action. 

Therefore,  we need to close the above gaps for the ODAD task shown in  Fig. \ref{fig:intro}. An emerging issue is how to effectively integrate action and appearance knowledge for ODAD. 
To this end,  we propose a novel Uncertainty-guided Appearance-motion Association Network (UAAN), which cleverly incorporates motion contexts into appearance-based object features for better detecting the actions among objects. Specifically, we first utilize the Faster R-CNN network \cite{ren2015faster} to extract the appearance-aware object features, and obtain the motion-aware object features by employing RoIAlign \cite{he2017mask} on the I3D network \cite{carreira2017quo}, which fully mines the object visual information. Then, we design two separate branches with the same architecture to reason the appearance-guided and  motion-guided object relations by graph convolutional networks (GCNs), respectively, so as to reason the inter-object interaction. Besides, we represent frame-level features by aggregating object features inside the frame. Finally, we design an  appearance-motion association attention module to integrate  appearance-aware object features and motion-aware object features for final detection. 

In summary, our contributions are as follows: 1) We  explore both   appearance- and motion-aware object information  for OOD action detection. Besides, we creatively propose a novel uncertainty-guided appearance-motion association network to reason inter-object interaction. 2) With the appearance and motion branch, we capture action-oriented and appearance-guided object relations via GCNs. By designing an appearance-motion association attention module, we integrate appearance and motion features from two branches for obtaining inter-object interaction. 3) Extensive experimental results on two challenging datasets show that our proposed model outperforms existing state-of-the-art approaches by a significant margin.

\section{Related Work}
\label{sec:related}
\noindent \textbf{Temporal action detection.}
Temporal action detection (TAD) \cite{zhao2017temporal,lin2017single,liu2022empirical} 
focuses on localizing action instances and classifying their categories in untrimmed videos \cite{richard2016temporal,liu2022end,zhao2023open}. 
Unlike traditional action detection \cite{jhuang2013towards,herath2017going,wang2013action,kong2022human} that focuses on identifying actions in videos without specifying their temporal boundaries, TAD involves precisely localizing the start and end times of the target actions  within an untrimmed video. 
Based on existing action detection models, OpenMax \cite{bendale2016towards} generates the action proposals, and employs OpenMax in testing to append the softmax scores with OOD actions.  DEAR \cite{BaoICCV2021} replaces the traditional cross-entropy loss for uncertainty quantification to detect OOD actions.  By predicting the locations, classifications with uncertainties, and actionness,  OpenTAL recognizes and localizes  all the actions simultaneously, and rejects OOD actions. 
TFE-DCN \cite{zhou2023temporal} designs a temporal feature enhancement dilated convolution network for weakly-supervised action detection. 
Recently, some  methods have been proposed based on the detection framework \cite{yeung2016end,de2016online,shou2016temporal}. For example, \cite{yeung2016end} aims to directly predict the boundary of target actions by designing  an end-to-end framework.
Although these TAD methods have achieved decent  success, they regard all the  actions in this video as belonging to some of the pre-defined classes.

\begin{figure}[t!]
    \centering
    \includegraphics[width=0.4\textwidth]{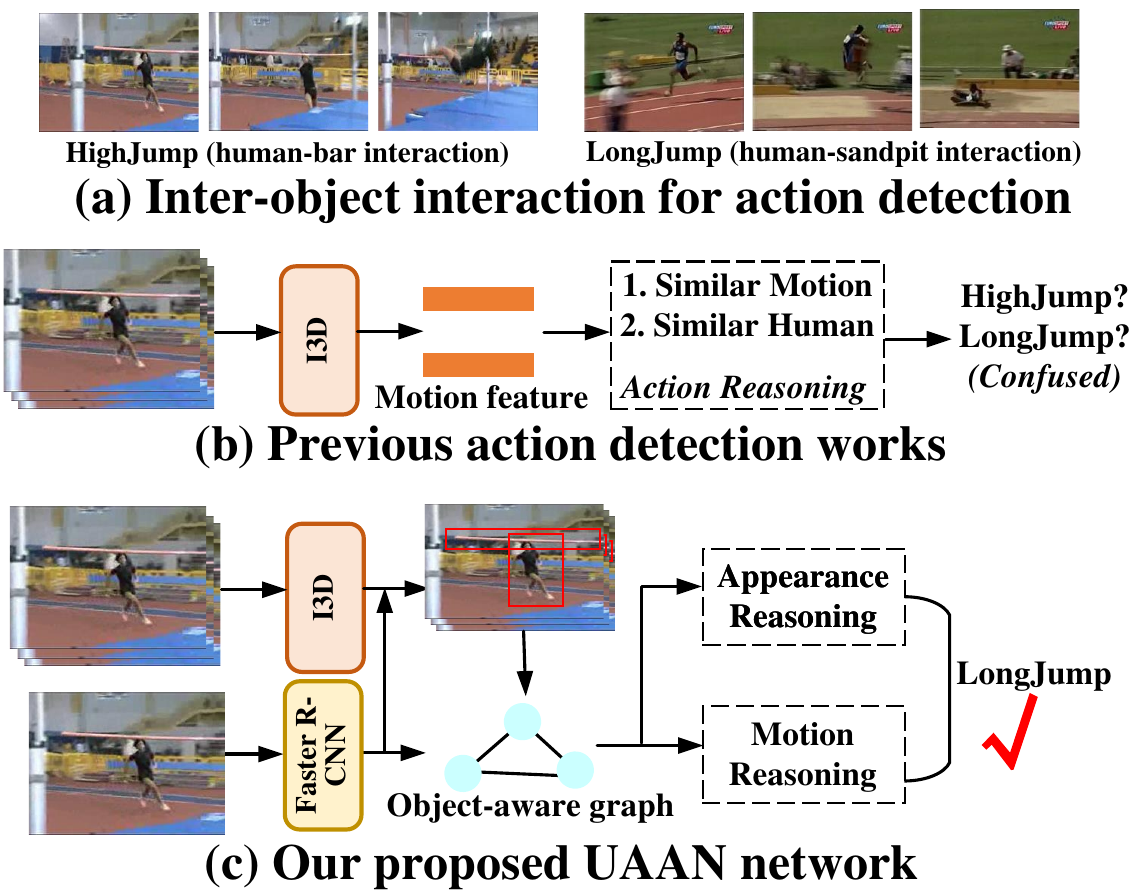}
     \vspace{-5pt}
    \caption{Our motivation. (a) Example of inter-object interaction for action detection. (b) Existing action detection
works only extracts frame-level motion information,  and fails to distinguish similar motions ``HighJump'' and ``LongJump''. (c) We construct an object-aware graph to reason the inter-object interaction from the appearance and motion perspectives.}
  \vspace{-15pt}
    \label{fig:motivation}
\end{figure}

Unfortunately, they only rely on the motion feature, which  captures the redundant background information and fails to perceive the fine-grained differences among video frames with high similarity. 
For example,  two real-world actions ``HighJump''  and ``LongJump'' belong to different action classes. As shown in Fig. \ref{fig:motivation}, the main difference between ``HighJump''  and ``LongJump'' is the appearance information:  ``high jump bar'' in ``HighJump'' and ``sandpit'' in ``LongJump''. Since they only model the motion features to capture the same motion ``jump'', they cannot correctly distinguish the local details of different objects (``high jump bar'' and ``sandpit'') in these frames, which will lead to wrong action detection results. Although some TAD methods \cite{yuan2017temporal,liu2022end} utilize the object detection technologies for action detection, these methods focus on single-object detection setting so that they only detect one object ``human'' and ignore other relevant objects. Thus, these methods cannot achieve satisfactory performance, where Fig. \ref{fig:vis_demo} illustrates two  qualitative results. Different from them, we creatively explore both appearance and motion features among different objects (\textit{e.g.,} ``high jump bar'' and ``sandpit'') for action detection. By  integrating appearance and motion features, we can correctly distinguish ``HighJump''  and ``LongJump''.

\noindent \textbf{Out-of-distribution detection.}
As a challenging computer vision task, out-of-distribution (OOD)  detection  targets to detect test samples from distributions that do not overlap with the training distribution. Previous OOD detection methods \cite{liang2018enhancing,liu2020energy,sun2021react,lee2018simple,mohseni2020self,vyas2018out,yu2019unsupervised,zaeemzadeh2021out,hsu2020generalized,ming2023exploit}  can be divided into four types: classification-based methods \cite{hendrycksbaseline,liang2018enhancing,lee2018hierarchical,lee2018training}, density-based methods \cite{kirichenko2020normalizing,serrainput}, distance-based methods \cite{techapanurak2020hyperparameter,lee2018simple} and reconstruction-based methods \cite{zhou2022rethinking,yang2022out}. 
1) Early OOD detection works refer to a classification framework, which utilizes the maximum softmax probability to determine the ID/OOD samples. 
2) To more explicitly model ID, density-based OOD detection methods are proposed to leverage the probabilistic models for OOD detection. These methods are under an operating assumption that OOD samples have low likelihoods whereas ID samples have  high likelihoods under the estimated density model. 
3) The distance-based OOD detection methods  are based on an intuitive idea that OOD samples should be relatively far away from the centroids of ID samples. 
under the Linear Discriminant Analysis assumption \cite{wallach2009rethinking}, and then utilizes the minimum Mahalanobis distance to all class centroids for OOD detection.
4) The reconstruction-based methods often leverage the encoder-decoder framework, which is trained on only ID samples and generates different outcomes for OOD detection. 

Different from previous OOD detection works that cannot handle  video datasets, our proposed method can detect OOD actions in the challenging  video dataset. Therefore, our method can be applied to more challenging scenarios than these image-based OOD detection methods.

\section{Our Proposed UAAN}
\label{sec:method}
\begin{figure*}[t!]
    \centering
    \includegraphics[width=\textwidth]{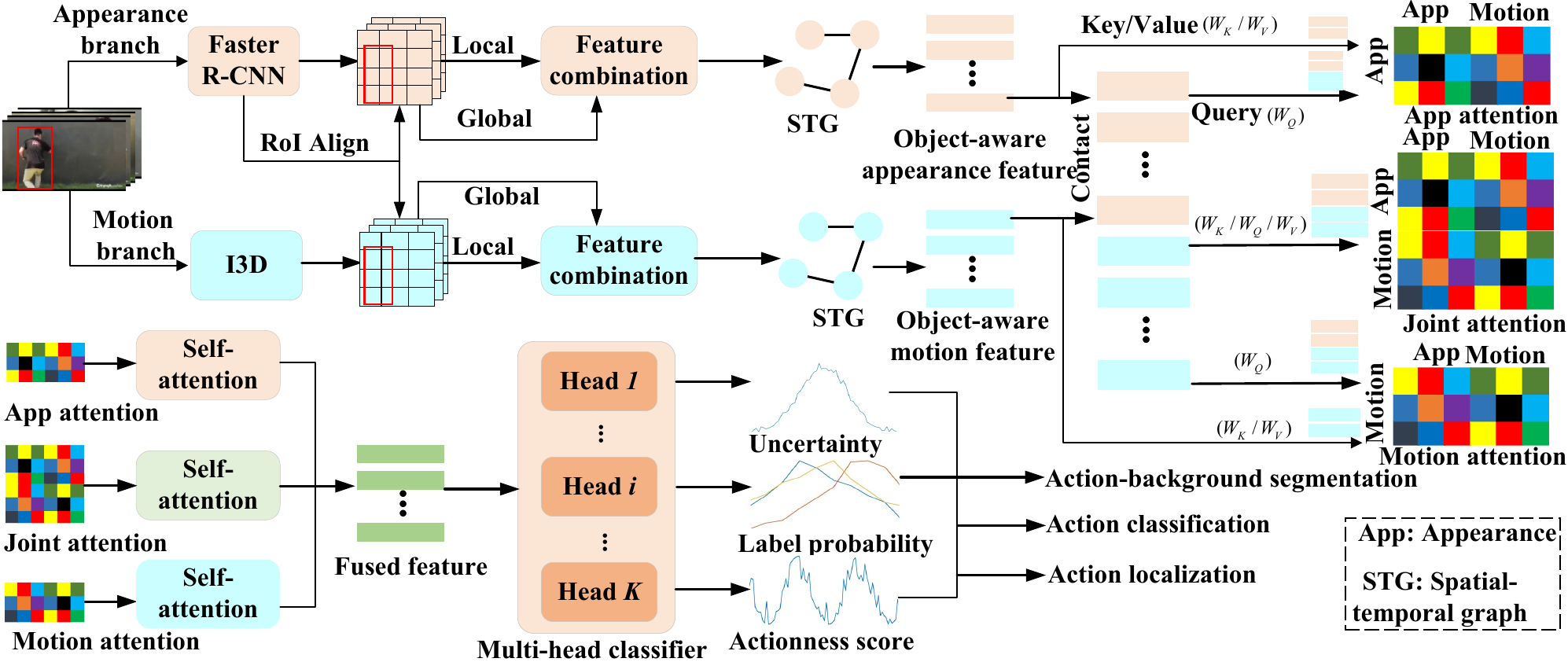}
    \caption{Overview of the proposed model for the challenging ODAD task. We first utilize video encoder (Faster R-CNN and I3D) to extract appearance- and motion-aware object features. Then, we design separate appearance and motion branches to reason the spatial-temporal interaction between different objects. Besides, we design an appearance-motion attention module to fully integrate the appearance and motion features for final detection. Best viewed in color.}
    \label{fig:pipeline_1}
\end{figure*}

Given an untrimmed video, the ODAD task aims to first localize all actions  with temporal boundary ($t_s,t_e$),  then recognize OOD actions, and finally classify all ID actions.  Given a training set $\{V_i,y_i,t_s^i,t_e^i\}_{i=1}^N$, where $N$ denotes the video number, $V_i$ denotes the $i$-th untrimmed video, $y_i\in \mathbb{R}^K$ is its multi-shot label denoting the action class that the action samples in $V_i$ belong to, and $K$ denotes the number of ID action classes. During training, the video data and the ID action labels are provided, while we cannot obtain the  OOD action labels. During testing, for any  untrimmed video $V^t$, we first distinguish if OOD actions are in $V^t$,  then classify each ID action in $V^t$ to one of the $K$ classes, and finally localize the boundaries of all the actions.

\subsection{Video Representation}  
Unlike previous action detection methods that only utilize the I3D network  to extract motion-aware features, we additionally consider extracting appearance-aware object features to explore inter-object interaction.

\noindent \textbf{Appearance representation.}
For a video with $T$ frames, we
utilize Faster R-CNN applying RoIAlign to extract the region of interest from a ResNet backbone \cite{he2016deep} and obtain $S$ objects. Thus, we can obtain $S\times T$ objects in each video. These object-aware appearance features are represented as $V_{local}^a=\{o_{t,k}^a, b_{t,k}\}_{t=1,k=1}^{t=T,k=S}$, where $o_{t,k}^a\in \mathbb{R}^d$ and $b_{t,k}\in \mathbb{R}^4$ indicate the local appearance feature and bounding box location of $k$-th object in the $t$-th frame, respectively. However, only the object-aware appearance features are not enough to fully understand the entire video. For example, we  cannot  distinguish ``HighJump'' and ``LongJump'' only based on a jumper. Thus, by another ResNet model with a linear layer,  we utilize another ResNet network to extract the frame-wise appearance representation $V_{global}^a\in \mathbb{R}^{T\times d}$.

\noindent \textbf{Motion representation.} 
We  extract the features of  video clips  by the last convolutional layer in an I3D network. For local motion features, we also  utilize RoIAlign to extract object-aware motion features: $V_{local}^m\!=\!\{o_{t,k}^m,\! b_{t,k}\}_{t=1,k=1}^{t=T,k=S}$, where $o_{t,k}^m\!\in\! \mathbb{R}^d$ and $b_{t,k}\!\in \!\mathbb{R}^4$. For global motion features, we  extract the clip-aware  features $V_{global}^m\!\in\! \mathbb{R}^{T\times d}$ by utilizing average pooling in the feature map and linear projection on the local motion features.

\noindent \textbf{Position representation.} 
In the challenging ODAD task, we consider  both spatial and temporal positions of each object for reasoning about object-wise action relations. Therefore,  a position encoding is added to object-level local features in both appearance and motion representations: $v_{t,k}^a=FFN([o_{t,k}^a;e^b;e^t])$ and $v_{t,k}^m=FFN([o_{t,k}^m;e^b;e^t])$, where $e^t$ is obtained by position encoding according to each frame's index, $e^b=FFN(b_{t,k})$, and $FFN(\cdot)$ is the feed-forward network. After the above process, we denote $\overline{V}_{local}^a=\{v_{t,k}^a\}_{t=1,k=1}^{t=T,k=S}$ and  $\overline{V}_{local}^m=\{v_{t,k}^m\}_{t=1,k=1}^{t=T,k=S}$.

Similarly, we add the position encoding $e^t$ into global appearance and motion features: $\overline{V}_{global}^a=FFN([V_{global}^a;e^t])$ and $\overline{V}_{global}^m=FFN([V_{global}^m;e^t])$. Thus, we  expand the above two global features from size $T\times d$ to $(T\times S)\times d$. To explore the context in objects, we concatenate object features with global features:
$F^a=FFN([\overline{V}_{local}^a;\overline{V}_{global}^a]), F^m=FFN([\overline{V}_{local}^m;\overline{V}_{global}^m])$, where $F^a=\{f_{t,k}^a\}_{t=1,k=1}^{t=T,k=S}\in \mathbb{R}^{(T\times S)\times d}$ are the final encoded object-level appearance features, and $F^m=\{f_{t,k}^m\}_{t=1,k=1}^{t=T,k=S}\in \mathbb{R}^{(T\times S)\times d}$ are the final encoded object-level motion features.

\subsection{Spatial-temporal Object Graph for Inter-object interaction}

We observe that utilizing inter-object interaction can distinguish motion-similar actions (\textit{e.g.}, ``HighJump'' and ``LongJump''). As shown in Fig. \ref{fig:motivation}, ``HighJump'' is the interaction between ``human'' and ``high jump bar'', while ``LongJump'' is the interaction between ``human'' and ``sandpit''. If we only use the common appearance information ``human'' and similar motion information ``jump'', we cannot correctly distinguish ``HighJump'' and ``LongJump''.

To explore the interaction between different objects for ODAD, we develop two separate branches based on appearance- and motion-aware object features to fully understand the video. In each branch, we reason about object relations by constructing a spatial-temporal graph.

\noindent \textbf{Object graph construction.}
Since the detected objects have both spatial interactivity and temporal continuity, we capture spatial-temporal relations by constructing object graphs in each branch, respectively. Since the two branches share the same architecture, we only describe the appearance branch’s components as an example to avoid redundancy.

In the appearance branch, we define the object-wise  appearance features $F^a=\{f_{t,k}^a\}_{t=1,k=1}^{t=T,k=S}$ including all objects in the whole video as nodes and build a fully-connected graph. Then, we adopt the graph convolution network (GCN) to extract the relation-aware object features via message propagation. In the object graph, we can obtain the adjacency matrix $A^a\in \mathbb{R}^{(T\times S)\times (T\times S)}$ to indicate the pairwise affinity between object features: $A^a=\sigma((F^aW_1^a)(F^aW_2^a)^\top)$,
where $W_1^a$ and $W_2^a$ are learnable parameters, and $\sigma(\cdot)$ is the softmax function. 

If two objects have strong semantic relationships, they will be highly correlated in the video, and they will have an edge with a high-affinity score in $A^a$. For deeper semantic reasoning, we introduce two-layer graph convolution with residual connection: $\hat{F}^a=LayerNorm(F^a+\tau(A^a\tau(A^aF^aW_3)W_4))$,
where $\tau(\cdot)$ denotes the Relu function, $W_3$ and $W_4$ are weight matrices of the GCN layers, and  $\hat{F}^a\in \mathbb{R}^{(T\times S)\times d}$ is the appearance-aware object features. Similarly, we can obtain  the motion-aware object features $\hat{F}^m\in \mathbb{R}^{(T\times S)\times d}$.

\noindent \textbf{Appearance and motion features association.}
After obtaining the object-wise features ($\hat{F}^a$ and $\hat{F}^m$), we integrate these features to  model inter-object interaction. Thus, we first concatenate $\hat{F}^a$ and $\hat{F}^m$: $U = \begin{bmatrix}\hat{F}^a\\\hat{F}^m\end{bmatrix}, \; U \in \mathbb{R}^{2(T\times S) \times d}$.
To effectively integrate different types of visual features, we first introduce the dot-product attention: $Att(W_Q, W_K, W_V)=\sigma(\frac{W_QW_K^\top}{\sqrt{d}})W_V$,
where $W_Q$, $W_K$ and $W_V$ are the query, key, and
value matrices in the dot-product attention, respectively.

To enhance different  features for inferring
action contexts, we introduce the attention function $Att(\cdot)$: 
$X^{a}\!=\!  LayerNorm(U+ Att(U,\hat{F}^{a}, \hat{F}^{a}))$, $X^{m}\!=\! LayerNorm(U+ Att(U, \hat{F}^{m}, \hat{F}^{m}))$, $X^{joint}\!=\! LayerNorm(U+ Att(U,U,U))$,
where $X\in \mathbb{R}^{(T\times S)\times d}$ is the enhanced features.
Then, we integrate these enhanced features by:

\begin{align}
\label{eq:integrate}
    X=Ave(X^{a},  X^{m}, X^{joint}), 
\end{align}
where  $Ave(\cdot)$ denotes the average pooling operation with $L2-$normalization, and $X=\{x_t\}_{t=1}^T$ is the fused feature.

\subsection{Uncertainty-guided OOD Action Detection}

\noindent \textbf{Action-background segmentation.}
Real-world videos naturally mix background and actions (ID and OOD actions), making it insufficient to recognize them only based on classification and uncertainty.
Therefore, we design an action detection module to indicate the likelihood of an frame belonging to an action. For convenience, we treat the action-background segmentation as a binary classification task.

To recognize class-agnostic actions from background frames, we follow \cite{AFSD_CVPR2021} to introduce the frame-aware actionness score: $\hat{a}_i\in[0,1]$. For any two frames $x_{t_1}$ and $x_{t_2}$ ($t_1\neq t_2$ and $x_{t_1},x_{t_2}\in\{1,\cdots,T\}$), they have three  relationships: 1) both $x_{t_1}$ and $x_{t_2}$ belong to the action; 2) both $x_{t_1}$ and $x_{t_2}$ are from the background; 3) one belongs to the action, while the other is from the background.  Therefore, we introduce the cosine similarity $\cos(x_{t_1},x_{t_2})$ and three thresholds ($\tau_{ bb},\tau_{aa},\tau_{dif}$) to evaluate their relationship. To supervise the action-background segmentation module by the annotated background frames and potential action frames, we introduce an affinity loss $\mathcal{L}_{ABS}$. Especially, $\mathcal{L}_{ABS}$ contains three significant parts ($\mathcal{L}^{\rm bg}$ based on two background frames, $\mathcal{L}^{\rm act}$ based on two action frames, and $\mathcal{L}^{\rm dif}$ based on the action-background pair).


 For $\mathcal{L}^{\rm bg}$, we use the online hard example mining strategy \cite{shrivastava2016training} to constrain two training background frames. 
We align two background frames into the same class by
the following loss: 

\begin{equation}
\mathcal{L}^{\rm bg}=\sum\nolimits_{a_{t_1}\leq a_{\tau}} \sum\nolimits_{a_{t_2}\leq a_{\tau}} \max[\tau_{ bb}-\cos(x_{t_1},x_{t_2}),0],
\end{equation}
where $a_{\tau}$ is an actionness threshold; $\tau_{bb}$ is the similarity threshold between two background frames from the background class. 
Also, we utilize a similar loss to align two action frames: 
\begin{equation}
\mathcal{L}^{\rm act}=\sum\nolimits_{a_{t_1}>a_{\tau}} \sum\nolimits_{a_{t_2}>a_{\tau}} \max[\tau_{ aa}-\cos(x_{t_1},x_{t_2}),0],
\end{equation}
where $\tau_{ aa}$ is the similarity threshold between two action frames from the action class. 
When $x_{t_1}$ and $x_{t_2}$ belong to action and background respectively, we can obtain the following loss: 
\begin{equation}
\mathcal{L}^{\rm dif}=\sum\nolimits_{a_{t_1}\leq a_{\tau}} \sum\nolimits_{a_{t_2}> a_{\tau}}  [\cos(x_{t_1},x_{t_2}) - \tau_{ dif},0],
\end{equation}
where $\tau_{dif}$ is the threshold to constrain the similarity between an action frame and a background frame. 
Based on the above three parts ($\mathcal{L}^{\rm bg}$, $\mathcal{L}^{\rm act}$ and $\mathcal{L}^{\rm dif}$), we can obtain the affinity loss $\mathcal{L}_{ABS}$ as follows:
\begin{equation}
\mathcal{L}_{ABS} = \frac{1}{3}(\mathcal{L}^{\rm bg} + \mathcal{L}^{\rm act} + \mathcal{L}^{\rm diff}).
\end{equation}

\noindent \textbf{Action classification.}
Different from previous OOD detection methods, we are required to  detect OOD actions in an untrimmed video. 
For an action class in real-world multimedia scenario, its predicted likelihood always follows a binomial distribution, and its conjugate prior is treated as a Beta distribution. 
We introduce evidential neural networks \cite{sensoy2018evidential} based on Beta distribution to jointly formulate the multi-class classification and uncertainty modeling. For the $i$-th action, we assume a Beta distribution $Beta(p_i|\alpha_i,\beta_i)$ over the action categorical probability $p_i\in [0,1]$, where  $\alpha_i$ and $\beta_i$ denote parameters to indicate the positive and negative evidence, respectively. In the feature space, the evidence evaluates actions closest to the predicted actions. Predicted labels should  be the same as positive evidence, but different from negative evidence. 

For action $i\in\{1,2,\cdots,K\}$, we can design a subjective opinion $w_i=(b_i,d_i,u_i,a_i)$ based on $\alpha_i$ and $\beta_i$, where belief $b_i\in [0,1]$, disbelief $d_i\in [0,1]$, uncertainty score $u_i\in [0,1]$, and $b_i+d_i+u_i=1$. 
For action $i$, the opinion $w_i$ is obtained by $\alpha_i$ and $\beta_i$: $b_i=\frac{\alpha_i-2}{\alpha_i+\beta_i},\quad d_i=\frac{\beta_i-2}{\alpha_i+\beta_i},\quad u_i=\frac{2}{\alpha_i+\beta_i}$.
The expected belief probability $p_i=b_i+a_i\cdot u_i$,
where $a_i$ is a base rate that denotes prior knowledge without commitment (neither agree nor disagree).

For  object $j$, its positive evidence is $\boldsymbol{\alpha}_j=s(h(\mathbf{x}_j;\boldsymbol{\theta}))+1$ and its negative evidence is $\boldsymbol{\beta}_j=s(h(\mathbf{x}_j;\boldsymbol{\theta}))+1$, where $\boldsymbol{\alpha}_j=[\alpha_{1j},\alpha_{2j},\cdots,\alpha_{Kj}]^\top$ and $\boldsymbol{\beta}_j=[\beta_{1j},\beta_{2j},\cdots,\beta_{Kj}]^\top$.
$\mathbf{x}_j$ means the  video, $h(\mathbf{x}_j;\boldsymbol{\theta})$ means the evidence vector and $\boldsymbol{\theta}$ mean parameters, $s(\cdot)$ means the evidence function (\textit{e.g.,} ReLU) to guarantee $\boldsymbol{\alpha}_j,\boldsymbol{\beta}_j \geq \mathbf{1}$.

For learning the above opinions, we introduce the Beta loss function by computing its Bayes risk for our action classification module. For the binary cross-entropy loss for each action $i$ over a batch of actors, we introduce the following loss:
\small
\begin{align}
\label{eq:beta-loss-1-actor}
    \mathcal{L}_{Beta} \nonumber
    =&\frac{1}{K}\sum\nolimits_{i=1}^K\sum\nolimits_{j=1}^S \int BC(y_{ij},p_{ij})Beta(p_{ij};\alpha_{ij},\beta_{ij})dp_{ij} \nonumber\\
    =&\frac{1}{K}\sum\nolimits_{i=1}^K \Big[y_{ij}\Big(\psi(\alpha_{ij}+\beta_{ij})-\psi(\alpha_{ij})\Big)\nonumber\\
    &+(1-y_{ij})\Big(\psi(\alpha_{ij}+\beta_{ij})-\psi(\beta_{ij})\Big)\Big],
\end{align}
where $K$ denotes the   action number,  $BC(\cdot)$ denotes the binary cross-entropy loss, and $\psi(\cdot)$ denotes the \textit{digamma} function. The log expectation of Beta distribution derives the last equality. For video $\mathbf{x}_j$,
$\mathbf{y}_j=[y_{1j},y_{2j},\cdots,y_{Kj}]\in\{0,1\}^K$ denotes its $K$-dimensional ground-truth action label.

\noindent \textbf{Action localization.} 
As shown in Fig. \ref{fig:pipeline_1}, we can predict each action proposal $l_i=(e_s^i, e_e^i)$ and a refined stage to predict the temporal offset $\delta_i=(\hat{e}_s,\hat{e}_e)$ by our model. Since the boundaries of pre-defined proposals are coarse, we employ a boundary regression loss for calibrating the boundary. We calculate a regression loss for every positive sample: 

\begin{align}
{\mathcal L}_{reg} = \frac{1}{{\mathcal N_p}}\sum\nolimits_i \mathcal{L}_1(e_s^i,\hat{e}_s^i)+ \mathcal{L}_1(e_e^i,\hat{e}_e^i),
\end{align}
where $\mathcal{L}_1$ denotes the smooth $l_1$ loss, $\mathcal{N}_p$ is the number of positive samples, $e_s$ and $e_e$ represent the truth error of proposal boundary, and $\hat{e}_s$ and $\hat{e}_e$ are the predicted errors.

To handle more complex scenarios, we additionally introduce the following Distance Intersection-over-Union (DIoU) loss \cite{zheng2020distance}: 

\begin{align}
{\mathcal L}_{DIoU} = \frac{1}{{\mathcal N}_p}\sum\nolimits_{i=1}^K (1-DIoU(l_i, \delta_i)).
\end{align}
Therefore, we can obtain the final localization loss:

\begin{align}
{\mathcal L}_{Local} = {\mathcal L}_{reg}+\gamma_0 {\mathcal L}_{DIoU},
\end{align}
where $\gamma_0$ is a parameter to balance  two losses.

\noindent \textbf{Training and inference.}
We train the proposed model by minimizing the following loss:

\begin{align}\label{final_loss}
{\mathcal L}_{final} =\gamma_1 {\mathcal L}_{ABS}  + \gamma_2{\mathcal L}_{Beta} + \gamma_3 {\mathcal L}_{Local},
\end{align}
where $\gamma_1$,  $\gamma_2$ and  $\gamma_3$ are  hyper-parameters to balance the importance between different losses.

During inference, we feed an untrimmed video into our proposed model, which  generates proposals comprising of a classification label $c_i$, an uncertainty score $u_i$, an actionness score $a_i$ and a predicted action location $l_i=(d_i^s,d_i^e)$. Therefore, by predefining  an uncertainty threshold $u_{\tau}$ and an actionness threshold $a_{\tau}$, we can directly determine whether the $i$-th action is an OOD action by the following process:

\begin{equation}
    P(x_i) = \left\{ 
    \begin{aligned}
    & \text{OOD action}, & \text{if} \; u_i > u_{\tau} {~\&~} a_i>a_{\tau}, \\
    & \text{ID action} (\hat{y}_i), & \text{if} \; u_i \leq u_{\tau} {~\&~} a_i>a_{\tau},\\
    & \text{Background}, & \text{otherwise}.
    \end{aligned}
    \right.
\label{eq:rule}
\end{equation}

\begin{table*}[!t]
\centering
\small
\caption{\textbf{Performance comparison (\%) vs. Different tIoU Thresholds}. Models trained on the THUMOS14 closed set are tested by including the OOD action classes from THUMOS14 and ActivityNet1.3, respectively. Results are averaged over the two dataset splits. }
\label{tab:thumos}
\setlength{\tabcolsep}{1.2mm}{
\begin{tabular}{l|l|cccccc|ccccccc}
\toprule
\multirow{2}{*}{Metrics} &\multirow{2}{*}{Methods} &\multicolumn{6}{c|}{THUMOS14 as the OOD set} &\multicolumn{4}{c}{ActivityNet1.3 as the OOD set} \\
\cline{3-12}
&&tIoU=0.3 &tIoU=0.4 &tIoU=0.5 &tIoU=0.6 &tIoU=0.7 &Mean &tIoU=0.5 &tIoU=0.75 &tIoU=0.95 &Mean \\
\hline
\multirow{6}{*}{AUROC($\uparrow$)}&SoftMax &54.70 &55.46 &56.41 &57.12 &57.11 &56.16 & 56.97 & 58.41 & 55.97 & 57.77 \\
&OpenMax \cite{bendale2016towards}&53.26 &52.10 &52.13 &51.89 &52.53 &52.38 & 51.24 & 52.39 & 49.13  & 51.59\\
&DEAR \cite{BaoICCV2021}&64.05 &64.27 &65.13 &66.21 &66.81 &65.29 & 62.82 & 66.23 & 67.92 & 65.69 \\
&OpenTAL  \cite{bao2022opental}&78.33 &79.04 &79.30 &79.40 &79.82 &79.18 & 82.97 & 83.21 & 83.38 & 83.22 \\
& TFE-DCN(+) \cite{zhou2023temporal}& 79.12& 79.56& 79.69& 78.34& 80.41& 79.85& 79.82& 82.73& 83.02& 83.13  \\
&\textbf{Our  UAAN}& \textbf{80.25}& \textbf{80.47}& \textbf{80.72}& \textbf{80.13}& \textbf{80.68}& \textbf{80.85}& \textbf{83.26} & \textbf{83.42}& \textbf{83.94}& \textbf{83.51}\\
\bottomrule
\multirow{6}{*}{AUPR($\uparrow$)}&SoftMax &31.85 &31.81 &31.11 &29.78 &27.99 &30.51 & 53.54 & 44.15 & 34.54 & 44.77 \\
&OpenMax \cite{bendale2016towards}&33.17 &31.61 &30.59 &29.15 &28.45 &30.60 & 54.88 & 48.37 & 40.07  & 48.48 \\
&DEAR \cite{BaoICCV2021}&40.05 &39.45 &38.05 &37.58 &36.35 &38.30 & 53.97 & 47.22 & 45.59 & 48.46 \\
&OpenTAL \cite{bao2022opental}&58.62 &59.40 &58.78 &57.54 &55.88 &58.04 & 80.41 & 74.20 &73.92 & 75.54\\
& TFE-DCN(+) \cite{zhou2023temporal}& 58.86& 58.72& 58.84& 57.35& 53.22& 58.80& 80.74& 70.28& 74.20& 76.08 \\
&\textbf{Our  UAAN}& \textbf{59.34} & \textbf{60.28}& \textbf{59.71}& \textbf{58.56}& \textbf{56.39}& \textbf{59.06}& \textbf{80.90}& \textbf{74.58}& \textbf{74.55}& \textbf{76.27}  \\
\bottomrule
\multirow{6}{*}{OSDR($\uparrow$)}&SoftMax &23.40 &25.19 &27.43 &29.97 &32.08 &27.61 & 27.63 & 33.73 & 31.59 & 32.01 \\
&OpenMax \cite{bendale2016towards}&13.66 &14.58 &15.91 &17.71 &20.41 &16.45 & 15.73 & 21.49 & 18.07 & 19.35\\
&DEAR \cite{BaoICCV2021}&36.26 &37.58 &39.16 &41.18 &42.99 &39.43 & 38.56 & 43.72 & 42.20 & 42.18 \\
&OpenTAL \cite{bao2022opental}& 42.91 &46.19 &49.50 &52.50 &56.78 &49.57 & 50.49 & 59.87 & 62.17 & 57.89 \\
& TFE-DCN(+) \cite{zhou2023temporal}& 43.05& 46.30& 48.72& 52.87& 57.02& 50.11& 50.83& 60.24& 63.01& 58.16\\
&\textbf{Our  UAAN}&\textbf{43.80} & \textbf{46.83}& \textbf{50.61}& \textbf{53.42}& \textbf{57.38}& \textbf{51.14} & \textbf{51.47} & \textbf{60.47}& \textbf{63.93}& \textbf{58.51}\\\bottomrule
\end{tabular}}
\end{table*}

\begin{table}[t!]
\centering
\caption{ODAD performance (\%), where we  set the tIoU threshold as 0.3 on  THUMOS14  and as 0.5 on the ActivityNet1.3. Models trained on  THUMOS14 closed set are tested on the OOD sets by including the OOD classes from THUMOS14 and ActivityNet1.3, respectively. mAP is provided as the reference of the localization results on THUMOS14 ID set.}
\label{tab:thumos_extra}
\vspace{-5pt}
\setlength{\tabcolsep}{1.2mm}{
\begin{tabular}{l|cc|ccccccc}
\toprule
\multirow{2}{*}{Methods} & \multicolumn{2}{c|}{FAR@95($\downarrow$) on different OOD sets}&\multirow{2}{*}{mAP($\uparrow$)}\\\cline{2-3}
&THUMOS14  &ActivityNet1.3   \\
\hline
OpenMax \cite{bendale2016towards}& 90.34 &  91.36 & 36.36 \\
SoftMax & 85.58&  85.05 & 55.81 \\
DEAR \cite{BaoICCV2021}& 81.42 & 84.01& 52.24 \\
OpenTAL \cite{bao2022opental}& 70.96 & 63.11& 55.02 \\
TFE-DCN(+) \cite{zhou2023temporal}&70.80& 62.54& 53.27\\
\textbf{Our  UAAN}&\textbf{69.43}&  \textbf{62.35}& \textbf{55.63}\\
\bottomrule
\end{tabular}}
\end{table}

\section{Experiments}
\label{sec:result}

\noindent \textbf{Datasets:}
For a fair comparison with existing works, we follow \cite{bao2022opental} to utilize two challenging and popular video datasets: THUMOS14~\cite{THUMOS14} and ActivityNet1.3~\cite{ANetCVPR2015}. THUMOS14 contains 200 validation videos and 212 testing videos from 20 labeled classes. ActivityNet1.3 has 19,994 videos with 200 action classes. 
We follow the former setting to split the dataset into training, validation and testing subsets by 2:1:1. 

\noindent \textbf{Evaluation metrics:}
To comprehensively evaluate the model performance,  we follow \cite{bao2022opental} to utilize the following evaluation metrics: mAP, AUROC, AUPR, OSDR, FAR@95, where a smaller FAR@95 value means better performance, while larger values for the other metrics denote better performance. 
Evaluation metrics with ``$\uparrow$'' indicate larger the better, while metrics with ``$\downarrow$'' mean  smaller the better.

\noindent \textbf{Implementation details.}
We leverage the two-stream I3D \cite{carreira2017quo} network as the backbone to extract video features, which is pre-trained on the Kinetics  dataset \cite{carreira2017quo}. For the THUMOS14 dataset, we treat 16 consecutive frames as a clip, a sliding window with a stride of 4 is used, and 1024-D features are extracted before the last fully connected layer. The two-stream features are further concatenated (2048-D). We utilize Adam optimizer with a learning rate of 0.0001 and with the mini-batch sizes of 16 and 64 for the THUMOS-14 and ActivityNet-v1.3 datasets, respectively. In the action-background segmentation module, we set $\tau_{bb}=0.3, \tau_{aa}=0.4, \tau_{dif}=0.4$.  Our codes are available in \href{https://github.com/AngeloFang/MIPR_UAAN}{Github}.

\subsection{Comparison with State-of-the-Art}

\noindent \textbf{Compared methods.}
Following \cite{bao2022opental}, we only compare the following state-of-the-art \textit{open-source} methods for  reproducibility: SoftMax, OpenMax \cite{bendale2016towards}, DEAR \cite{BaoICCV2021}, OpenTAL \cite{bao2022opental} and TFE-DCN \cite{zhou2023temporal}.
Since TFE-DCN  cannot be directly used for the ODAD task, we embed TFE-DCN into OpenTAL by replacing the action detection module in  OpenTAL. To distinguish it from the original model, we denote the embed model as TFE-DCN(+).
Different from them, we explore the inter-object interaction for better understanding video actions.

\noindent \textbf{Performance comparison.}
Tables~\ref{tab:thumos} and \ref{tab:thumos_extra} report  performance comparison, where all the methods are tested using both the THUMOS14 unknown splits and the ActivityNet1.3 disjoint subset. For fair evaluation, we report all performance comparisons by averaging the results of each evaluation metric over the three THUMOS14 splits.  The best results are in \textbf{bold}. 
Obviously, our UAAN surpasses all the compared methods by significant margins on all the metrics.

\begin{figure}[t!]
\centering
\includegraphics[width=0.24\textwidth]{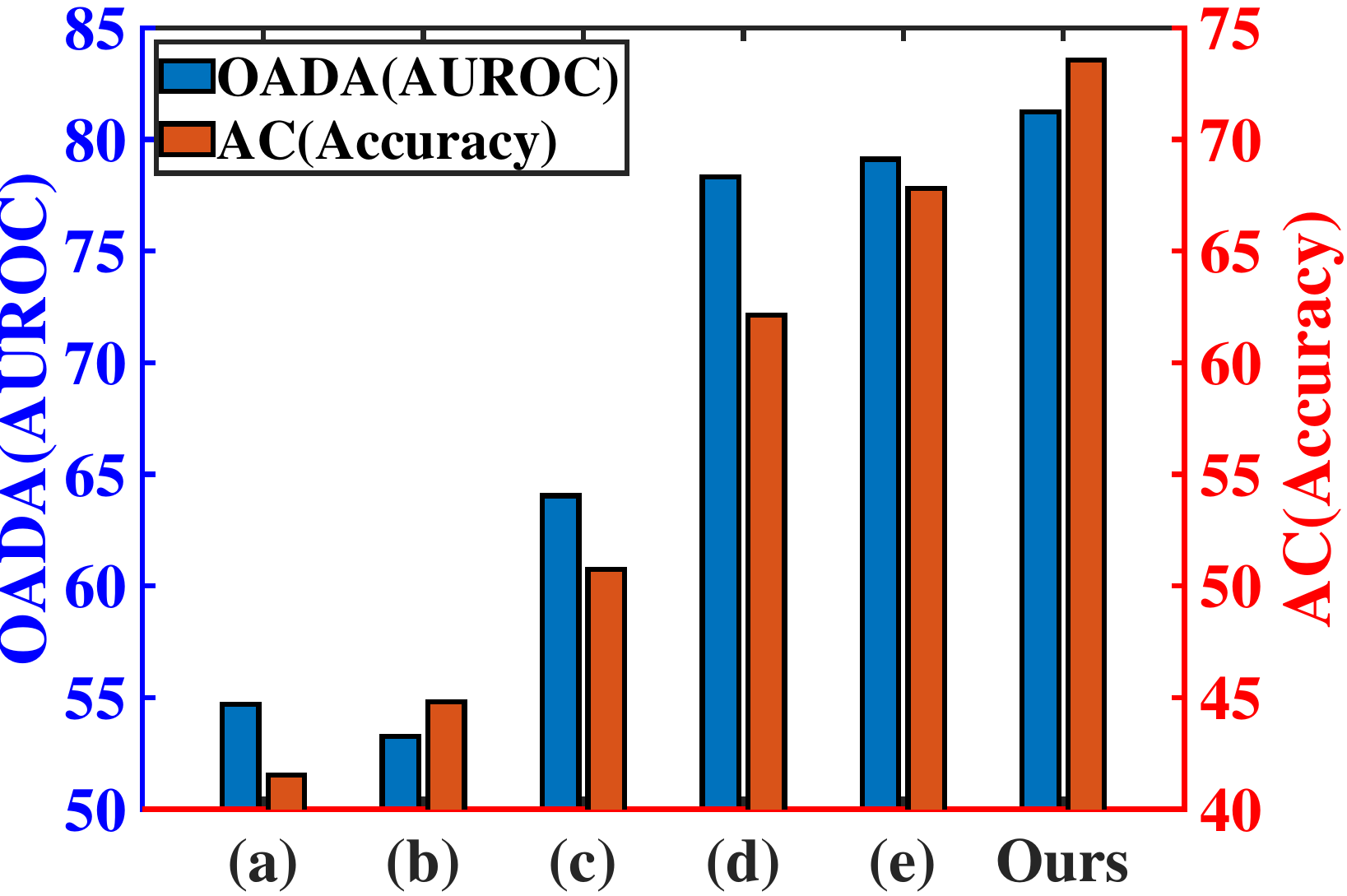}
\hspace{-0.07in}
\includegraphics[width=0.24\textwidth]{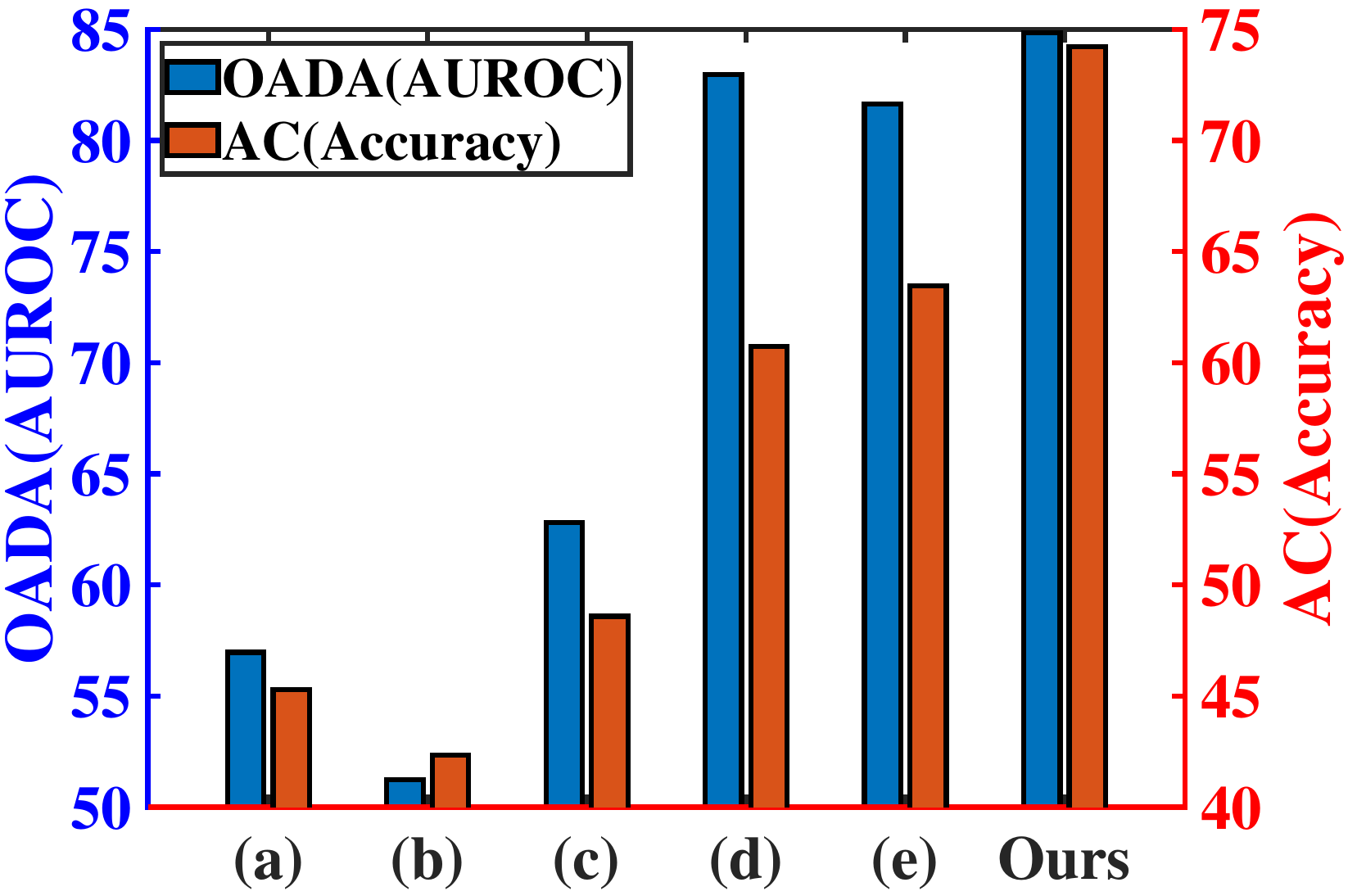}
\caption{ Performance comparison for ID action classification and ODAD on THUMOS14 (left) and ActivityNet1.3 (right), where ``ODAD(AUROC)'' means ``AUROC for ODAD'' and ``AC(Accuracy)'' means ``Classification accuracy for ID action classification''. (a) is SoftMax, (b) is OpenMax \cite{bendale2016towards}, (c) is DEAR \cite{BaoICCV2021}, (d) is OpenTAL \cite{bao2022opental}, and (e) is TFE-DCN(+) \cite{zhou2023temporal}. }
\label{fig:close_vs_open}
\end{figure}

\noindent \textbf{ID action classification
and ODAD.} We further compare the performance of both ID action classification
and OOD action detection in Fig. \ref{fig:close_vs_open}. Obviously, our UAAN significantly outperforms all compared methods over two evaluation metrics (AUROC and Accuracy), showing the effectiveness of our UAAN for both ID action classification
and ODAD.

\begin{table}[t!]
\caption{Main ablation study on  THUMOS14, where we remove each key individual component to investigate its effectiveness. ``OSG'' is ``object-aware spatial-temporal graph'', ``MAFA'' is ``motion and appearance features association'', and ``UOAD'' is ``uncertainty-guided OOD action detection''.}
\scalebox{1}{
\setlength{\tabcolsep}{1.1mm}{
\begin{tabular}{ccc|cccccccccccccccccc}
\hline
{OSG}&{MAFA}&{UOAD}& FAR@95($\downarrow$) & AUROC($\uparrow$) & AUPR($\uparrow$) & OSDR($\uparrow$)\\\hline
\XSolidBrush & \CheckmarkBold & \CheckmarkBold & 73.47& 75.88& 57.12& 39.80 \\
\CheckmarkBold & \XSolidBrush & \CheckmarkBold & 71.75& 74.08& 55.93& 40.02  \\
\CheckmarkBold & \CheckmarkBold & \XSolidBrush & 72.82& 74.43& 56.20& 41.38 \\\hline
\CheckmarkBold &\CheckmarkBold &\CheckmarkBold &\textbf{69.43}& \textbf{80.25}& \textbf{59.34}& \textbf{43.80}\\ \hline
\end{tabular}}}
\label{tab:main_ablation_thu}
\end{table}
\subsection{Ablation Study on  THUMOS14}

\noindent \textbf{Main ablation study.} We perform exhaustive ablation studies to analyze the effectiveness of each individual component. Table \ref{tab:main_ablation_thu} shows the main ablation results on  THUMOS14. Specifically, we remove each module to evaluate its significance.
The result shows that our full model has precision improvement compared with each ablation model, which manifests that each above component provides a positive contribution. 
Compared with the first ablation model, our full model improves performance by 4.04\% in terms of ``FAR@95''. It is because our object-aware spatial-temporal graph can fully mine the inter-object interaction for action reasoning. For the second ablation model, our full model beats it by 6.17\% over ``AUROC''. The main reason is that we can integrate both appearance and motion visual information of each object for action detection. Besides,  our full model outperforms the third ablation model by 5.82\% over ``AUROC''. The significant improvement is because we can segment action and background by $\mathcal{L}_{ABS}$ and leverage inter-object interaction for action classification by $\mathcal{L}_{Beta}$.

\begin{figure}[t!]
\centering
\includegraphics[width=0.24\textwidth]{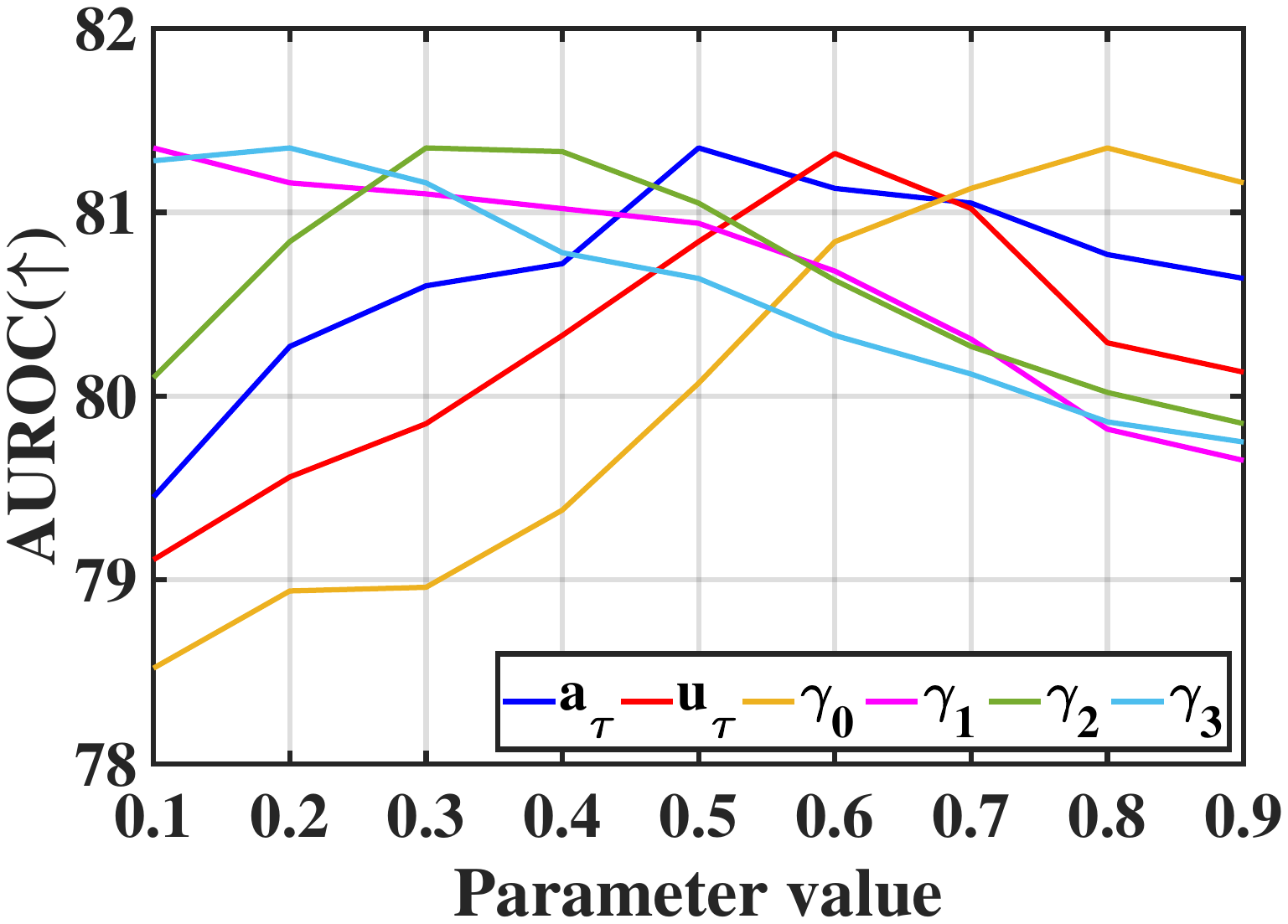}
\hspace{-0.07in}
\includegraphics[width=0.24\textwidth]{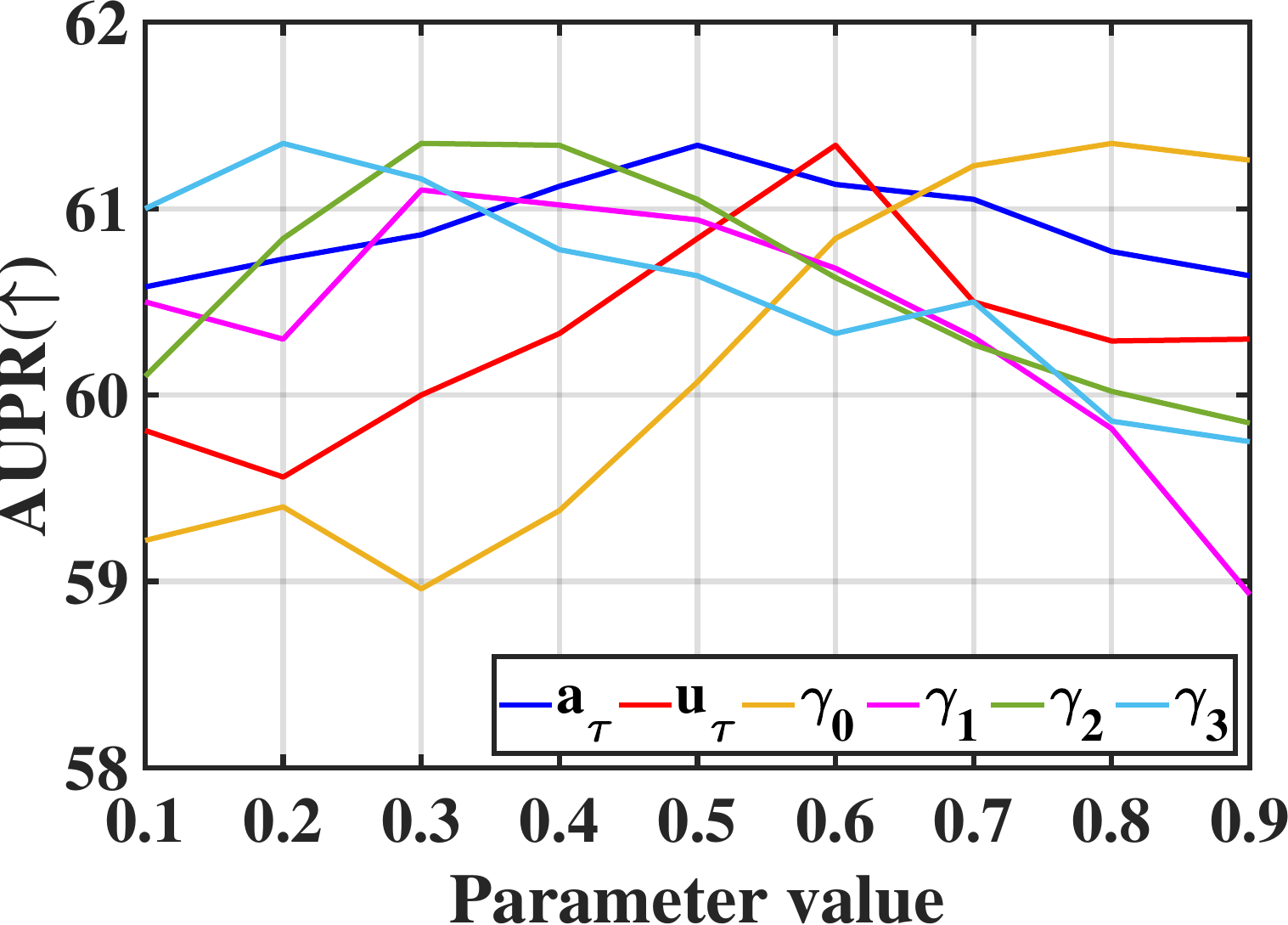}
\caption{Analysis on  hyper-parameters. Best viewed in color.}
\label{fig:canshu}
\end{figure}

\noindent \textbf{Hyper-parameters analysis.} 
We conduct the ablation studies on the hyper-parameters $a_{\tau},u_{\tau},\gamma_0,\gamma_1,\gamma_2,\gamma_3$ in Fig.~\ref{fig:canshu}. For each ablation study, we change  one  hyper-parameter by fixing the others.
We can obtain the best performance when  $a_{\tau}=0.5,u_{\tau}=0.6,\gamma_0=0.8,\gamma_1=0.15,\gamma_2=0.3,\gamma_3=0.2$.

\section{Conclusion}
In this paper, we target a challenging task: ODAD, which is an extension of OOD detection and action detection. To this end, we propose the novel UAAN method to explores both  appearance features and motion contexts to reason spatial-temporal inter-object interaction. 
Experimental results confirm the effectiveness of our proposed UAAN.

{
 \bibliographystyle{IEEEtran}
\bibliography{main}}

\end{document}